\documentclass[sigconf]{acmart}
\usepackage{multirow}
\usepackage{multicol}
\usepackage{graphicx}
\usepackage{subfigure}
\usepackage[utf8]{inputenc}
\usepackage[russian, english]{babel}
\babeltags{ru = russian}

\renewcommand\footnotetextcopyrightpermission[1]{} 

\def\BibTeX{{\rm B\kern-.05em{\sc i\kern-.025em b}\kern-.08emT\kern-.1667em\lower.7ex\hbox{E}\kern-.125emX}}
    
\setcopyright{rightsretained}
\acmConference[SIGIR eCom'20]{ACM SIGIR Workshop on eCommerce}{July 30, 2020}{Virtual Event, China}
\acmYear{2020}
\copyrightyear{2020}
\makeatletter
\renewcommand\@formatdoi[1]{\ignorespaces}
\makeatother
\acmISBN{}

\begin{document}

%
\title{Exploiting Neural Query Translation into \\Cross Lingual Information Retrieval}
%
\author{Liang Yao ~~~~~~~ Baosong Yang ~~~~~~~ Haibo Zhang}
\authornote{Corresponding author.}
\author{ Weihua Luo ~~~~~~~ Boxing Chen} 
\affiliation{
\institution{Alibaba Group}
\city{Hangzhou, China}
}
\email{ {yaoliang.yl, yangbaosong.ybs, zhanhui.zhb, weihua.luowh, boxing.cbx}@alibaba-inc.com }  


%
\renewcommand{\shortauthors}{Yao et al.}

%
\begin{abstract}
As a crucial role in cross-language information retrieval (CLIR), query translation has three main challenges: 1) the adequacy of translation; 2) the lack of in-domain parallel training data; and 3) the requisite of low latency. 
To this end, existing CLIR systems mainly exploit statistical-based machine translation (SMT) rather than the advanced neural machine translation (NMT), limiting the further improvements on both translation and retrieval quality. 
In this paper, we investigate how to exploit neural query translation model into CLIR system. 
Specifically, we propose a novel data augmentation method that extracts query translation pairs according to user clickthrough data, thus to alleviate the problem of domain-adaptation in NMT. Then, we introduce an asynchronous strategy which is able to leverage the advantages of the real-time in SMT and the veracity in NMT. 
Experimental results reveal that the proposed approach yields better retrieval quality than strong baselines and can be well applied into a real-world CLIR system, i.e. Aliexpress e-Commerce search engine.\footnote{Readers can examine and test their cases on our website: \url{https://aliexpress.com}.}


\end{abstract}


%
\maketitle
\section{Introduction}
Query translation serves as an important step in cross-language information retrieval (CLIR)\cite{gao2006statistical}. Generally, CLIR system first translates user query to the language in terms of its index database, the translation quality therefore significantly affects the retrieval results\cite{wu2010study}. 

Recently, neural machine translation (NMT) have shown their superiorities in a variety of translation tasks and even yielded human-level performance \cite{vaswani2017attention}.
However, existing CLIR systems still apply traditional translation models, e.g. bilingual dictionaries or statistical machine translation systems (SMT) \cite{koehn2009statistical,och2002discriminative}, which to some extent restrict the further development of CLIR context. 
The inapplicability of NMT in CLIR system is mainly lies in three concerns:
\begin{itemize}
    \item[{\bf C1}:] Query is usually the distillation of user intention. 
    Contrast to SMT, NMT lacks a mechanism to guarantee all the source words to be translated, resulting in fluent but inadequate translations~\cite{he2016improved,martindale2019identifying,bi2020constraint}.
    Several researchers may doubt on the accuracy and the completeness of a translation generated by NMT model~\cite{yarmohammadi2019robust,sarwar2019multi,rubino2020effect,wan2020unsupervised}.
    \item[{\bf C2}:] Recent studies have proven that a well-performed NMT model depends on extensive language resources~\cite{popel2018training,ott2018scaling,Yang2020improve}. The lack of in-domain query pairs may obstruct its training, thus weakening its performance \cite{yao2020domain}.
    \item[{\bf C3}:] As a component in CLIR, query translation has to be real-time, while the advanced NMT model built with multiple network layers are fail to cope with the requirement on processing speed. 
\end{itemize}
To the best of our knowledge,  there is few real-world CLIR system exploits NMT and explores its effectiveness. To this end, we aim at resolving the problems list above, thus to make the advanced techniques in terms of NMT contribute the CLIR community.

In response to the first concern ({\bf C1}), we conduct experiments on a widely used multi-lingual search task--CLEF to empirically assess the retrieval quality with respect to SMT and NMT models (Section 5.2).  The experimental results shows that queries translated by NMT model perform better on retrieval tasks than that generated by SMT model. This not only dispels the doubt on the inablity of NMT on query translation, our further qualitative analysis but also demonstrates that NMT model can better handle word disambiguation and rare word translation than its SMT counterpart.

Then, we try to further improve neural query translation by tackling the domain-adaptation problem ({\bf C2}) using a novel data augmentation approach. 
Users across the world issue multilingual queries to the search engines of a website everyday, which form large-scale cross-lingual clickthrough data.  
Intuitively, when the translation of a query leads the user to click details and even make purchases on recalled items, we attribute the translation pair possesses high quality and enables the expection of users. 
With the help of such an automatic and low cost  quality estimation approach, our model can acquire a large amount of in-domain query translation pairs derived from user behaviors. These training samples can improve the performance of our translation model, of which outputs are further examined by users, fulfilling a positive cycle.  



Finally, in order to overcome the problem of latency ({\bf C3}) and deploy NMT model into the real-world e-commerce system, we design a new architecture that features synchronous call to SMT model and asynchronous call to NMT model. 
Specifically, we employ a cache to store query translation pairs. If the user query fails to hit translation memories, our system requests the real-time SMT engine to return the translation candidates, and asynchronously updates the cache with the translation of NMT model. 
In this way, our system can jointly profits from the low latency of SMT and the high quality offered by NMT.  

We assess the effectiveness of our approaches upon a real-word e-Commerce search engine--Aliexpress. Experimental results demonstrate that the mined data significantly increases the word coverage of our training data, thus to improve the quality of query translation. Moreover, the new architecture makes its average latency comparable to that of SMT (on the magnitude of 10ms), while raises the  retrieval accuracy with the help of NMT. 

\section{Background}


\subsection{Neural Machine Translation}


Neural machine translation (NMT) \cite{bahdanau2014neural} is a recently proposed approach to machine translation which builds a single neural network that takes a source sentence
\begin{math} x=(x_{1},...,x_{T_x}) \end{math}
as an input and generates its translation
\begin{math} y=(y_{1},...,y_{T_y}) \end{math}
, where
\begin{math} x_{t} \end{math}
and
\begin{math} y_{t^{'}} \end{math}
are source and target symbols. Ever since the integration of attention, NMT systems have seen remarkable improvement on translation quality. Most commonly, an attentional NMT consists of three components: (a) an encoder which computes a representation for each source sequence; (b) a decoder which generates one target symbol at a time, shown in  Eq.1 ; (c) the attention mechanism which computes a weighted global context with respect to the source and all the generated target symbols.

\begin{equation}
\log p(y|x)= \sum_{t=1}^{T_y} \log p(y_t|y_{t-1},x)
\end{equation}

Given N training sentence pairs 
\begin{math} {(x^i,y^i)\dots(x^n,y^n)\dots(x^N,y^N)} \end{math},
Maximum Likelihood Estimation (MLE) is usually accepted to optimize the model, and training objective is defined as:

\begin{equation}
L_{MLE}=-\sum_{n=1}^{N}\log  p(y^{n}|x^{n})
              =-\sum_{n=1}^{N}\sum_{t=1}^{T_{y}}\log p(y_{t}^n|y_{t-1}^n,x^n)
\end{equation}

Among all the encoder-decoder models, the recently proposed Transformer \cite{vaswani2017attention} architecture achieves the best translation quality so far. In this paper, we introduce the most advanced Transformer model architecture into the query translation, which greatly reduces the ambiguity of translation, and improves the quality of retrieval. 

The Transformer architecture relies on a self-attention mechanism \cite{lin2017structured} to calculate the representation of the source and target side sentences, removing all recurrent or convolutional operations found in the previous methods. Each token is directly connected to any other token in the same sentence via self-attention. The hidden state in the Transformer encoder is calculated based on all hidden states of the previous layer. The hidden state 
\begin{math} h_{t}^{i} \end{math}
in a self-attention network is calculated as in Eq.3.

\begin{figure}[h]
  \centering
  \includegraphics[width=0.80\linewidth]{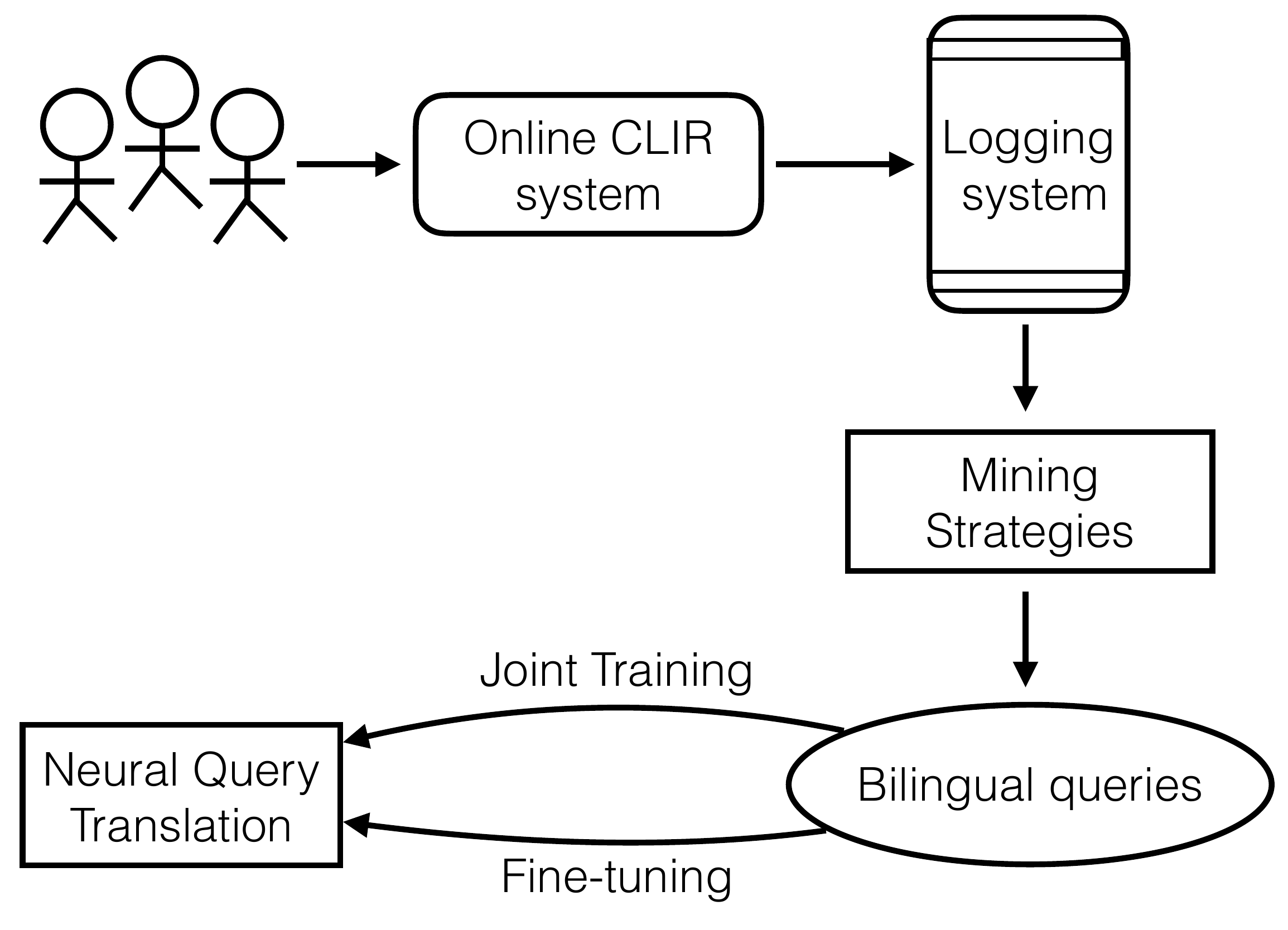}
  \caption{The framework of optimizing neural query translation for CLIR using user clickthrough data.}
  \Description{The whole process of optimizing CLIR with data of user behavior}
\end{figure}

\begin{equation}
h_{t}^{i}=h_{t-1}^{i}+F(self{-}attention(h_{t-1}^{i}))
\end{equation}

where \begin{math} F \end{math} represents a feed-forward network with layer normalization and ReLU as the activation function. The decoder additionally has a multi-head attention over the encoder hidden states. For more details, refer to Vaswani \cite{vaswani2017attention}.

\section{Data Augmentation with Clickthrough Record}



In this section, we first introduce the overview of our proposed method. Then, we describe how to record clickthrough data and how to mine in-domain query translation pairs. At last, the method to exploit neural query translation with these mined data will be introduced.

\subsection{Overview}

Often, the query fed to any information retrieval system can be rather “unnatural”, which is very short or in wrong word order or repetitive in meaning \cite{nikoulina2012adaptation}.  It is difficult for general-purpose machine translation systems which are usually trained on a corpus of standard parallel sentences to deal with such query phrases properly. This problem can be seen as a problem of domain adaptation, where the domain is query \cite{Shadi16}. To our knowledge, no suitable corpus of parallel queries is available to train an adapted machine translation system. Although small corpora of parallel queries can be obtained (eg. CLEF tracks) or manually created, it is insufficient for a commercial query translation system.

Intuitively, millions of users across the world issue queries to the search engines of Aliexpress e-commerce website in various languages everyday, which forms large-scale and live cross-lingual clickthrough data. If a large number of users click on one or more items in the search results page of a multilingual query to check the details, or even make a purchase, we believe that the user's expectations have been retained and that the translation of the multilingual query is relatively good. We can utilize the user behavior in the clickthrough data to mine query translation pairs. Such user clickthrough data can be obtained in large quantities and recorded at a very low cost.

Based on this assumption, we propose a novel method to optimize query translation using NMT enhanced with in-domain bilingual translation pairs mined from clickthrough data of cross-lingual search engine. The whole process is presented in Figure 1: a mass of users search, browse, click or purchase on cross-language search websites everyday. These data will be recorded in the logging system. After excavating valuable features from massive user behaviors, we then acquire bilingual query translation pairs through these features. We then apply these bilingual queries to adapt an NMT system through either data augmentation or fine-tuning method.

\subsection{Recording Clickthrough Data}
In the scenario of cross-language search on Aliexpress e-commerce website, a complete search process is as follows. After a user inputs a multilingual query, it will first be translated into English. The search engine will then search this English query in the English index database. Then, a search result list in the same language as the original query will be sent to the user who can decide whether to click on any items in the search result or not at all. Such a search process can be formulated as below: where \begin{math} u \end{math} stands for user, \begin{math} q \end{math} is the multilingual query, \begin{math} t \end{math} indicates the translation of the query and \begin{math} c \end{math} records the click result of user, respectively.

\begin{equation}
d=<u,q,t,c>
\end{equation}

We can collect massive data like this in real cross-language search scenario. With the definition of metadata, in the next section we will explain how to mine bilingual queries from massive search logging data.

\subsection{Mining Query Translation Pairs}
When the user expectation in real cross-language search scenario has been retained, the translation of the multilingual query is to be relatively good. Mining query translation hidden in the user clickthrough data is based on the following two criteria.

\paragraph{\bf The first criterion:} When a user clicks on one or a few items in the search result page of a multilingual query to check the details and even makes purchases, we consider the translation of the multilingual query is to be relatively good, and that the user expectation has been preserved. The translation conveys the true expectation of user. When the translation carries critical errors, such as, mistranslation, omission, addition and other translation errors,  the search result will be less relevant and the user will click less. Based on this observation, we can define a conversion rate (\begin{math}CTR\end{math}) of a pair of one multilingual query and its translation \begin{math}<q,t>\end{math}, as a ratio of detail unique view (\begin{math}Duv\end{math}) to list unique view (\begin{math}Luv\end{math}). "Detail" is the description page of product. "List" is the list page of search result of a query.  \begin{math}Luv_{<q,t>}\end{math} is defined in Eq.5 as the number of users who have all searched the same multilingual query \begin{math}q\end{math}. \begin{math}Duv_{<q,t>}\end{math} in Eq.6 is the sum of users who have also clicked through on at least one items in the search result page.

\begin{equation}
Luv_{<q,t>}=\sum_{u}\Theta \left \langle u,q,t \right \rangle
\end{equation}

\begin{equation}
Duv_{<q,t>}=\sum_{u}\Theta \left \langle u,q,t \right \rangle\cdot  \Psi \left ( c \right )
\end{equation} 

where \begin{math} \Theta  \left \langle u,q,t \right \rangle \end{math} in Eq.7 is an indicator that a user \begin{math} u \end{math} issues a multilingual query \begin{math} q \end{math} with translation \begin{math} t \end{math} to the search engine and  \begin{math} \Psi
\end{math} in Eq.8 represents whether or not the user clicks the items on the search result page.

\begin{equation}
\Theta \left \langle u,q,t \right \rangle = 1
\end{equation}

\begin{equation}
\Psi \left ( c \right )= \left\{\begin{matrix}
1,c\geq 1 & \\ 
0,c=0 & 
\end{matrix}\right.
\end{equation}

Once \begin{math}Luv_{<q,t>}\end{math} and \begin{math}Duv_{<q,t>}\end{math} are computed, we can work out the conversion rate, i.e., \begin{math}CTR\end{math}, as Eq.9,

\begin{equation}
CTR_{<q,t>}=\frac{Duv_{<q,t>}}{Luv_{<q,t>}}
\end{equation}

Obviously, higher conversion rate \begin{math} <q,t> \end{math} indicates more reliable translation; otherwise, there might be some problems with the translation or search ranking algorithm. Therefore, translation quality is reliable when the conversion rate is higher than certain threshold value. We may filter out unreliable search query and its translation by following Eq.10 below.

\begin{equation}
CTR_{<q,t>} \geq \eta 
\end{equation}

\paragraph{\bf The second criterion:} Sometimes, click data may be unreliable, especially when a query rarely appears in searches. In this case, whether the user makes clicks or not is extremely random, even when the translation is perfect. We exclude the \begin{math} <q,t> \end{math} pairs when the original query is rare by forcing \begin{math}Luv_{<q,t>}\end{math} higher than a threshold.

\begin{equation}
Luv_{<q,t>} \geq \chi 
\end{equation}

Only those queries \begin{math}<q,t>\end{math} met two conditions above will be mined and constitute final in-domain bilingual data. Loads of users browse and click everyday on CLIR system, which helps us in choosing and labeling translations of queries. Therefore, we can continuously collect a number of free in-domain bilingual data.

\subsection{Training Strategy}


In this part, we exploit neural query translation with these mined in-domain translation pairs. Two methods are proposed to enhance the neural translation model, one of which is data augmentation and the other is fine-tuning. 







\paragraph{\bf Joint Training (JT):} The mined bilingual queries are  mixed with common training data, which are altogether used to train an NMT system from scratch. This is the simplest and most direct approach of applying mined bilingual queries to improve the translation quality of NMT.

\begin{figure}[h]
  \centering
  \includegraphics[width=0.93\linewidth]{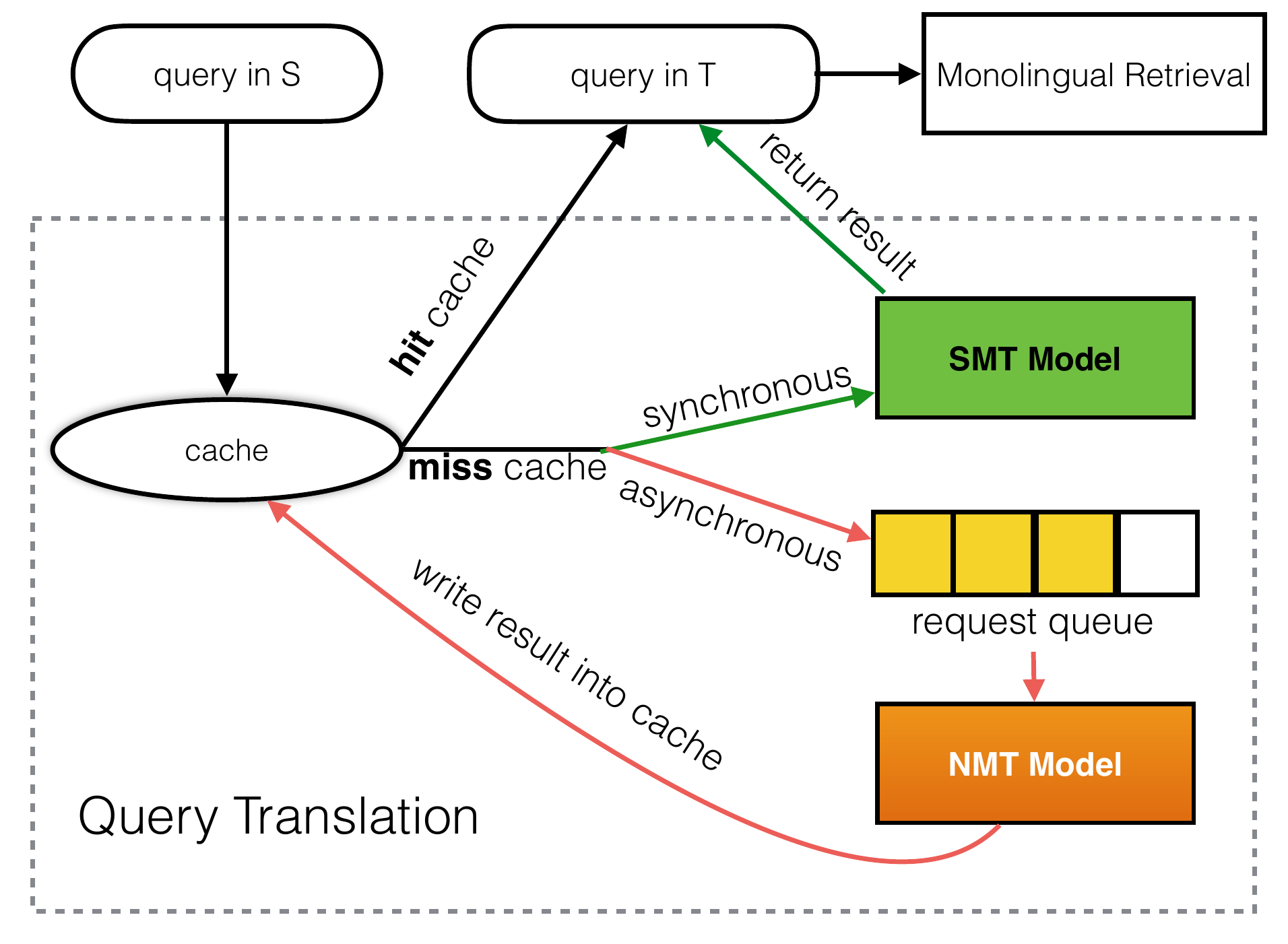}
  \caption{The architecture of the proposed CLIR system. A query in source language (S) from a user will be first translated by the  translation module, then retrieve from the commodity library with the well-translated query in target language (T). 
  In the query translation module, we use a cache scheme to bridge asynchronous call and synchronous call. If there is a hit, the model will return the translation result directly. Otherwise, it will synchronously request the SMT engine to acquire the translation result and push the query into a queue being asynchronously translated by NMT.}
  \Description{The architecture of proposed CLIR system.}
\end{figure}

\paragraph{\bf Fine-tuning (FT):} Fine-tuning is a fast and simple adaptation method, which has demonstrated substantial improvements in various neural network based models including NMT \cite{sennrich2015improving,gulcehre2015using,chu2017empirical}. To fine tune an NMT, we first train a baseline NMT system with common training data until convergence. Then, we replace the common training data with the mined bilingual queries and continue training until another convergence.


\section{Asynchronous Translation Architecture}

In the previous section, we have deliberately explained the details of our proposed method. Although a large breakthrough has achieved in translation quality, the requirement of low latency in the actual commercial CLIR scenario prevents the NMT technology from directly applying to the query translation system. It is very difficult for practical application in the commercial CLIR scenario because the translation speed of NMT is about 10 times slower than SMT. Through data analysis on Aliexpress e-commerce website, we found out that there is only a small amount of queries repeatedly assessed by a large number of users every day. The query repetitive rate is up to over 90\%.


Given the low latency requirement and high repetitive rate, we propose a new system architecture with the combination of synchronous call to SMT and asynchronous call to NMT for query translation for CLIR. As shown in Figure 2, the architecture is able to maintain the latency at the same millisecond speed of SMT, and complete the user retrieval with a guaranteed quality. The architecture has been broadly tested in the real business scenario of Aliexpress e-commerce website. The process of our architecture is mainly as follows: a query in S from a multilingual user will be first translated by query translation module, then retrieve from the commodity library with the well-translated query in T, and display the retrieved commodity to the user with the language set by the website after document translation. After obtaining the commodity information, the user can browse, click, add to wish-list or purchase the commodity. 

In the Query Translation module, we devise a new system architecture that profits on both low latency of SMT and quality offered by NMT.
We use a cache scheme to bridge asynchronous call and synchronous call. When a query is requested, it will take priority access to the cache which is only used to store the NMT results, if there is a hit, it will return the translation result directly. Otherwise, it will synchronously request the SMT engine to acquire the translation result at a millisecond level, and return the result directly without writing into the cache. At the same time, it will asynchronously request to add to the queue to be translated by NMT, after translation is completed, the translation result will be stored in the cache. So in the next time when another same query is requested, it can return the NMT result at the millisecond-level. The architecture proposed is capable of meeting over 90\% of queries using NMT for the translation result every day in the real business scenario.
\begin{figure*}[htbp]
\centering
\subfigure[The distribution of queries with different $ Luv $]{
\begin{minipage}[t]{0.45\linewidth}
\centering
\includegraphics[width=3in]{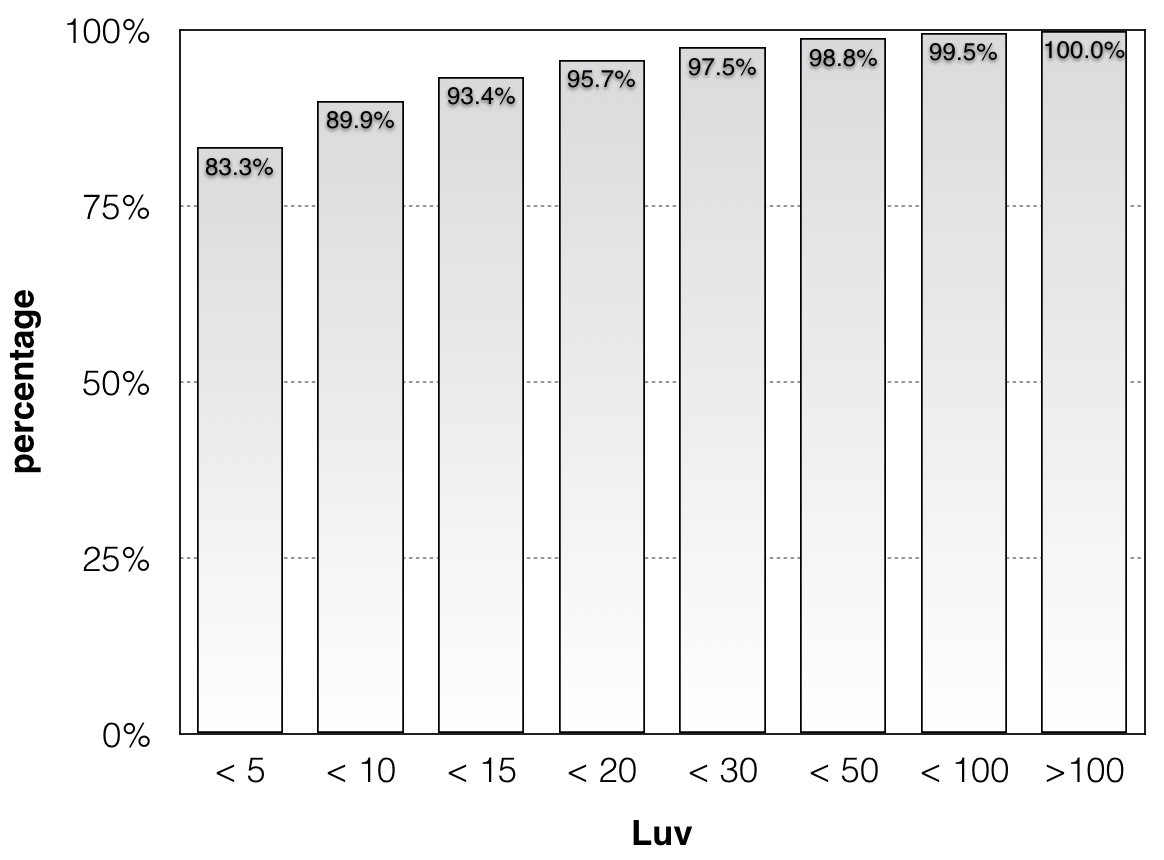}
\end{minipage}
}
\quad\quad\quad
\subfigure[The distribution of queries with different $ CTR $]{
\begin{minipage}[t]{0.45\linewidth}
\centering
\includegraphics[width=3in]{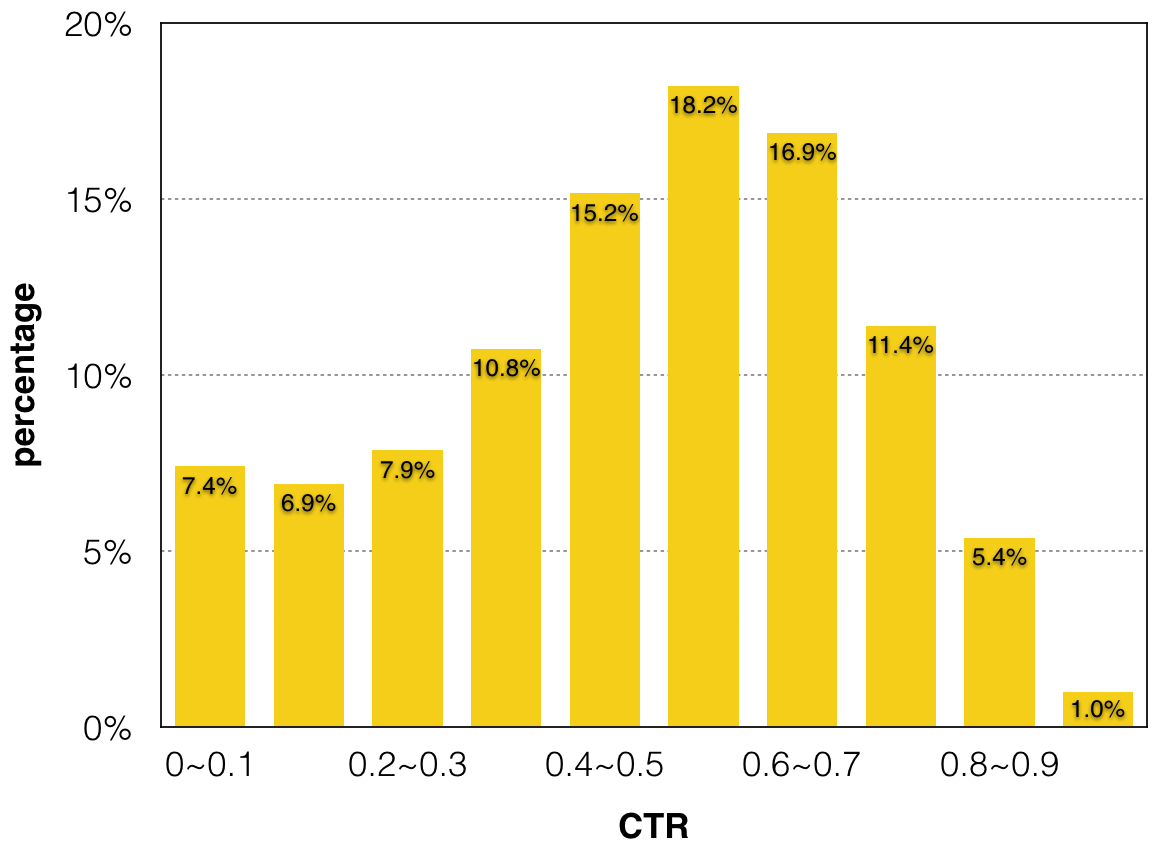}
\end{minipage}
}
\caption{ (a) The relationship between the ratio of queries and $Luv$  (a) ; and $CTR$ (b). X-axis indicates the threshold of $Luv$ and $CTR$, respectively. Y-axis  presents the ratio of the number of queries to the whole.
For instance, the interval of $0\leq CTR\leq$ reveals that there exists $7.41\%$ queries with $CTR $ being less than $0.1$.}
\end{figure*}

\section{EXPERIMENTS}

In this section, we conducted experiments to answer the following three questions.
\begin{itemize}
    \item [1.]  Is the NMT model better than SMT for query translation to improve retrieval quality in CLIR task?
    \item [2.]  Does the NMT enhanced with mined bilingual in-domain query translation pairs from the user clickthrough data provide improvements in retrieval quality?
    \item [3.]  Can the proposed system architecture provide online services in the real scenario ?
\end{itemize}

\subsection{Setup}
\subsubsection{Data} \

To answer the first question, we undergo the multilingual ad-hoc search task (IRTask4) of CLEF eHealth Consumer Health Search Task 2018 \cite{saleh2018cuni,suominen2018overview} which is similar to the IR tasks in the previous years (2013–2017). In this task, the monolingual English queries were translated by experts into Czech(CZ), French(FR) and German(DE), and the participants are asked to design a search system to retrieve relevant documents to these queries from the English document collection. Two indexes of the document collection in the CLEF 2018 consumer health search task are provided, in the first one, documents are stemmed and a stop-word list is used, while no preprocessing is done in the second index. The collection contains 5,560,074 documents. We use the first index in our experiments. The development set includes queries from the IR task of 2014 (50 queries), 2015 (66 queries), 2016 (300 queries) and 2017 (300 queries), respectively. After removing duplications, there are 416 queries remaining. The test set includes 50 queries from CLEF 2018. The training corpora for NMT contains the UFAL data set and the WMT2018 data set. UFAL Medical Corpus is a collection of parallel corpora aiming at more reliable machine translation of medical texts. UFAL provides in-domain medical data and out-of-domain generic text. WMT2018 provides out-of-domain generic text. The Table 1 shows the composition of the training data.

\begin{table}
  \label{tab:freq}
  \caption{Statistics of the training data we used in CLEF eHealth 2018 Task4.}
  \begin{tabular}{|c|c|c|c|}
    \hline
    Language&In-domain&Out-of-domain&Total\\
    \hline
    CZ-EN&1,145,493&57,279,765&58,425,258\\
    \hline
    DE-EN&3,036,581&73,518,668&76,555,249\\
    \hline
    FR-EN&3,418,951&88,213,528&91,632,479\\
    \hline
  \end{tabular}
\end{table}

To answer the second question, we collect user clickthrough data from Aliexpress e-commerce website in October 2018 and select data according to the two dimensions of \begin{math}CTR\end{math} and select data according to Eq.5-11. We use the CLIR system that is serving millions of users through the Aliexpress e-commerce website every day. In terms of \begin{math}Luv\end{math}, more than 80\% of all queries are used by less than 5 people, as summarised in Figure 3a. In consideration for credible statistics, we choose \begin{math}Luv\end{math} equal or greater than 15 and calculate the \begin{math}CTR\end{math} distribution of the chosen queries, as shown in Figure 3b. It can be seen that the average value of \begin{math}CTR\end{math} is greater than 0.5. Setting the \begin{math}CTR\end{math} threshold at 0.7 and 0.3, we finally selected 100k \begin{math}<q,t>\end{math} pairs as the additional data for training NMT. The main bulk of training data for NMT comes from the WMT2016 Russian-to-English news translation task as well as Russian-to-English e-commerce data. The size of the development set and the test set is 2000, which are extracted from the search log of the Aliexpress e-commerce website in 2017 and produced by human translators. The overall data composition is shown in Table 2.

\subsubsection{Systems} \

\begin{table}
  \label{tab:freq}
  \caption{Statistics of the samples we used for evaluating the proposed data augmentation method. "Top" clickthrough data consists of the queries whose $CTR$ is greater than $0.7$, while "Bottom" clickthrough data denotes the group of queries with  $CTR$ being less than $0.3$. }
  \begin{tabular}{|c|c|c|}
    \hline
    Type&Source&Size\\
    \hline
    \multirow{4}*{Training} & Top clickthrough data & 100k\\
        \cline{2-3}
		~ & Bottom clickthrough data & 100k\\
        \cline{2-3}
		~ & WMT 2016 data & 2.4M\\
        \cline{2-3}
		~ & E-commerce bilingual data & 14M\\
    \hline
    Validation&E-commerce query with human translation&2k\\
    \hline
    Test&E-commerce query with human translation&2k\\
    \hline
  \end{tabular}
\end{table}
\begin{table*}
  \label{tab:freq}
  \caption{Experimental Results on CLEF eHealth 2018 Task4 which consists of Czech-English, German-English and French-English cross-lingual search tasks. $\dagger$ represents NMT is significantly better than baseline SMT system.}
  \begin{tabular}{cccccccccc}
    \hline
     & \multicolumn{3}{c}{Czech} & \multicolumn{3}{c}{German} & \multicolumn{3}{c}{French} \\
    System & P@10 & MAP & NDCG@10 & P@10 & MAP & NDCG@10 & P@10 & MAP & NDCG@10 \\
    \hline
    Monolingual & 71.20 & 27.24 & 60.46 & 71.20 & 27.24 & 60.46 & 71.20 & 27.24 & 60.46 \\
    CUNI & 63.00 & 18.92 & 51.04 & 62.00 & 18.88 & 53.19 & 66.80 & 22.0 & 56.33 \\
    SMT & 64.40 & 21.46 & 53.28 & 65.40 & 21.61 & 55.85 & 62.60 & 20.67 & 53.05 \\
    SMT+BPE & 64.00 & 21.46 & 53.97 & 63.20 & 22.02 & 53.91 & 62.00 & 20.53 & 52.41 \\
    \hline
    \textbf{NMT} & \textbf{66.60} & \textbf{22.83} & \begin{math}\textbf{55.85}^{\dagger}\end{math} & \begin{math}\textbf{69.00}^{\dagger}\end{math} & \textbf{23.21} & \textbf{58.36} & \begin{math}\textbf{66.80}^{\dagger}\end{math} & \begin{math}\textbf{22.73}^{\dagger}\end{math} & \begin{math}\textbf{57.10}^{\dagger}\end{math} \\
    \hline
  \end{tabular}
\end{table*}

\begin{table*}
  \label{tab:freq}
  \caption{Case study on translation examples output by SMT and NMT. ``SRC'' and ``REF'' denote the source query and its translation reference, respectively.  As seen, NMT translates a Czech phrase ``dítě rýma'' to ``runny nose'', while SMT generates ``fever'', leading to completely irrelevant search results. Considering the German example, NMT is able to translate the word "bewältigungsstrategien" with complex morphology. }
  \begin{tabular}{c|l|r|r|r}
    \hline
    Task  & \multicolumn{1}{c|}{Translation Example} & P@10 & MAP & NDCG@10 \\
    \hline
   \multirow{4}{*}{Czech$\Rightarrow$English} & SRC: čtyřměsíční dítě rýma & & & \\
    &REF: four month infant running nose & 70.00 & 22.12 & 54.53 \\
    &SMT: four-month-old baby fever & 0.00 & 0.00 & 0.00 \\
    &NMT: a four-month-old runny nose & 40.00 & 3.66 & 35.56 \\
    \hline
    \multirow{4}{*}{German$\Rightarrow$English} &SRC: angst bewältigungsstrategien & & & \\
    &REF: anxiety coping skills & 40.00 & 14.88 & 33.33 \\
    &SMT: fear bewältigungsstrategien & 0.00 & 0.00 & 0.00 \\
    &NMT: fear management strategies & 30.00 & 3.58 & 13.44 \\
    \hline
  \end{tabular}
\end{table*}
\textbf{NMT: }We adopt the Transformer model with base setting as defined in \cite{vaswani2017attention} for experiments. We use BPE\cite{sennrich2015neural} to preprocess the source and target texts, forming source-side and target-side dictionaries with 30,000 and 30,000 tokens, respectively. The optimizer used for the MLE training is Adam \cite{kingma2014adam} with initial learning rate at 0.003. We follow the same learning rate schedule as in \cite{vaswani2017attention}. During training, roughly 4,096 source tokens and 4,096 target tokens are paired in one mini batch. Each model is trained using 8 NVIDIA Tesla P100 GPUs. For inference, we use beam search with width 6. We run each setting for at least 5 times and report the averaged score on test set. The test set score is chosen via the best configuration based on the validation set. 

{\bf SMT:} We use Moses \cite{koehn2007moses} to train the baseline SMT system. The phrase length is set to 5 and the n-gram of the language model \cite{stolcke2002srilm} is 3. SMT systems are tuned toward BLEU \cite{papineni2002bleu} using MERT \cite{och2003minimum}. 

{\bf Retrieval System:} On the CLEF eHealth 2018 task4, we use the Terrier \cite{ounis2006terrier,ounis2005terrier} with Dirichlet model as the retrieval system. On the experiments to verify the impact of the user clickthrough data, we use the online information retrieval system of Aliexpress e-commerce website.

\subsection{NMT vs SMT for Query Translation}

We performed the experiments for Czech-English, German-English and French-English language pairs of the multilingual ad-hoc search task (IRTask4) of CLEF eHealth Consumer Health Search Task 2018. The following systems are compared to show the impact of NMT . 
\begin{itemize}
\item  Monolingual: retrieval using the manually translated English queries provided with the corpus.
\item  CUNI: retrieval using 1-best translation of Khresmoi \cite{hanbury2011khresmoi} SMT system without re-ranking and query expansion. CUNI \cite{saleh2018cuni} is the best system submitted on this task.
\item  SMT: retrieval using 1-best translation of Moses phrase-based SMT system without re-ranking and query expansion.
\item  NMT: retrieval using 1-best translation of Transformer system without re-ranking and query expansion.
\end{itemize}

\begin{figure}[t]
  \centering
  \includegraphics[width=0.85\linewidth]{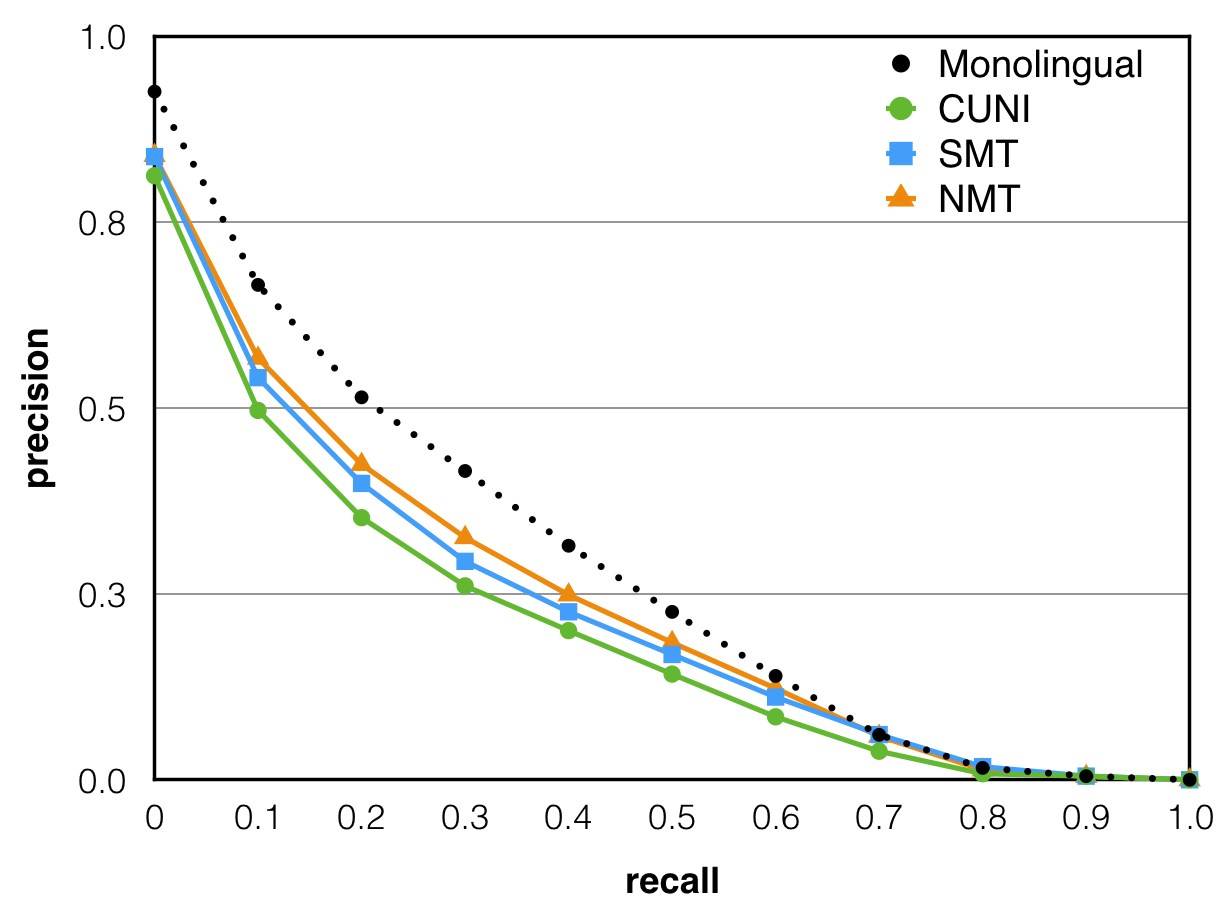}
  \caption{P-R curve of Czech-English in CLEF 2018}
  \Description{P-R curve of Czech-English in CLEF 2018}
\end{figure}
\begin{table*}
  \label{tab:freq}
  \caption{
  Effectiveness of the proposed data augmentation method on general domain training set. The user clickthrough data is collected from our Aliexpress website. 
  We treat the NMT model trained on WMT 2016 news data as baseline.  JT and FT stand for exploiting the augmented data with joint training and fine-tuning, respectively. ``coverage'' indicates the proportion of words in the test set that appear in the training set. $\dagger$ represents the system is significantly better than baseline.}
  \begin{tabular}{ccccccc}
    \hline
    Train data & System & BLEU-4 & P@10 & MAP & NDCG@10 & Coverage \\
    \hline
    & Monolingual & & 71.33 & 73.77 & 70.37 & \\
    \hline
    WMT2016 & Baseline & 27.23 & 42.66 & 44.16 & 42.14 & 85.22\% \\
    WMT2016 + Top clickthrough data & + JT & 34.15 & \begin{math}\textbf{60.33}^{\dagger}\end{math}  & \begin{math}\textbf{61.67}^{\dagger}\end{math}  & \begin{math}\textbf{59.84}^{\dagger}\end{math}  & 91.70\% \\
    WMT2016 + Top clickthrough data & + FT & \textbf{34.40} & \begin{math}\textbf{57.00}^{\dagger}\end{math} & 57.30 & \begin{math}\textbf{56.86}^{\dagger}\end{math} & 91.70\% \\
    WMT2016 + Bottom clickthrough data & + JT & 32.57 & 53.66 & 56.53 & 52.61 & 92.97\% \\
    WMT2016 + Bottom clickthrough data & + FT & 31.06 & 55.33 & 56.06 & 55.06 & 92.97\% \\
    \hline
  \end{tabular}
\end{table*}

\begin{table*}
  \label{tab:freq}
  \caption{
   Effectiveness of the proposed data augmentation method on e-Commerce domain training set. The baseline is trained with our in-house e-commerce data. As seen, the proposed methods can further improve the translation model that trained with in-domain data.}
  \begin{tabular}{ccccccc}
    \hline
    Train data & System & BLEU-4 & P@10 & MAP & NDCG@10 & Coverage \\
    \hline
    & Monolingual & & 71.33 & 73.77 & 70.37 & \\
    \hline
    E-commerce & Baseline	& 32.40	& 59.66	& 61.04	& 59.16	& 91.04\% \\
	E-commerce + Top clickthrough data &+ JT& 34.78 & \begin{math}\textbf{66.33}\end{math}  & \begin{math}\textbf{67.08}\end{math}  & \begin{math}\textbf{66.03}\end{math}  & 93.33\% \\
    E-commerce + Top clickthrough data &+ FT& \textbf{35.48} & 64.66 & 66.40 & 64.09	 & 93.33\% \\
	E-commerce + Bottom clickthrough data &+ JT& 33.78 & 59.66 & 61.65 & 59.01	 & 94.96\% \\
	E-commerce + Bottom clickthrough data &+ FT& 34.54 & 59.00 & 60.98 & 58.28	 & 94.96\% \\
    \hline
  \end{tabular}
\end{table*}
The experimental results are shown in Table 3. On the P@10, MAP and NDCG@10 indicators, our NMT system has surpassed the baseline SMT and CUNI system. In addition, we show the P-R curve of the Czech-to-English task in Figure 4. The results of German-to-English and French-to-English tasks are similar to that of the Czech-to-English task. For the significance test between SMT and NMT system, we use Wilcoxon test \cite{fix1955significance} with \begin{math}\alpha = 0.05\end{math} \cite{hull1993using}, which is used to compare two independent samples with paired data taken from the same distribution. \begin{math}\dagger\end{math} represents NMT is significantly better than SMT system. We also try to integrate SMT with BPE. However, the results show that BPE is less effective when applied to SMT. From these experimental results, we can conclude that NMT can better preserve the query's intention than SMT in query translation.

In order to understand why NMT performs better than SMT in the CLIR task, we analyse the translation results of the test set from Czech-to-English and German-to-English, respectively. As shown in Table 4, the second row is a Czech-English test example. The third row is an example of German-English test. For the Czech phrase "dítě rýma", NMT translates it into "runny nose", while SMT, however, translate this phrase into "fever", leading to completely irrelevant search results. From the point of alleviating translation ambiguity, NMT performs better than SMT even when the query is short and lacks rich context. For the German word "bewältigungsstrategien" with complicated morphology, it is difficult to cover the word in the training corpus, so SMT can not translate the word, but NMT uses the sub-word technology to split the word and is able to translate the sub-word units as a sequence. In conclusion, NMT has the natural advantages in disambiguation and translating rare and out-of-vocabulary words over SMT.

Overall, we believe NMT is a better choice for CLIR for the first question. Therefore, we will verify the effect of mined bilingual query translation pairs from the user clickthrough data on the system of NMT. 

\subsection{Effect of User Clickthrough Data}

To answer the second question, we carry out two sets of experiments, where the base training data is of either news articles or e-commerce data and the systems are trained using various combinations of all training data. Transformer is the NMT model and we use the search engine of Aliexpress e-commerce website for document retrieval. The experimental results are shown in Table 5 and 6, where JT stands for joint training and FT for fine-tuning, respectively.

As we can see from the tables, the bilingual query translation pairs mined from the user clickthrough data can significantly improve translation quality and retrieval quality in either general or e-commerce experiments. Such mined bilingual pairs improve the word coverage of the test set in the training set, which is usually beneficial for translation quality. The word coverage indicates the proportion of words in the test set that appear in the training set. Although the word coverage of the test set in the top clickthrough data is lower than that in the bottom clickthrough data, better retrieval quality is observed when the top clickthrough data is used for training the NMT systems. This is mainly because the translations of the queries in the top clickthrough data is more trustworthy and the search engine can find more and better matching results. Nonetheless, the bottom clickthrough data is still somewhat useful in improving retrieval quality. We find that some bottom clickthrough data can also lead to improvement in retrieval quality. The main reason could be either some problems in the ranking process of search engine or that bottom clickthrough data also contains high quality \begin{math}<q,t>\end{math} pairs.

In both experiment sets, the best system in terms of the quality of query translation is fine-tuned with the top clickthrough data in both sets of experiments, while the best retrieval quality improvement is observed for the NMT system trained with augmented data of the base training data and the top clickthrough data. The observation indicates that the fine-tuning method may overfit the system towards the top clickthrough data, which is actually harmful for retrieval quality. This is also in line with previous reports that improving translation quality does not guarantee higher relevance of retrieved documents to the original query in CLIR systems \cite{pecina2014adaptation,Fujii09}.
\begin{table}[t]
  \label{tab:freq}
  \caption{Comparison of performance among different systems. ``Proportion'' presents the proportion of our system calls on SMT and NMT model. }
  \begin{tabular}{|c|c|c|}
    \hline
    System&Average Latency& Proportion \\
    \hline
    SMT & 9.77ms & 100\% SMT \\
    \hline
    NMT & 148.43ms & 100\% NMT \\
    \hline
    Our architecture & 12.87ms & 10\% SMT + 90\% NMT \\
    \hline
  \end{tabular}
\end{table}
\subsection{Performance of System Architecture}
To answer the third question, we collect the user search queries in Russian for one month from Aliexpress e-commerce website and randomly select 30 million queries  as test data without deduplication. Since we have mainly innovated in the query translation module among the entire CLIR system, we only test the performance of this module. We compare the following systems SMT, NMT, and our proposed architecture. SMT system is deployed on a five-core CPU server and NMT on an eight-core GPU server of the Tesla P100.

The experimental results are shown in Table 7. The proposed architecture basically achieves the performance of SMT in terms of average latency and has the ability to provide online services. Since the latency of NMT is too high, it is difficult to provide services to users in real business scenarios.  Although in recent researches\cite{zhang2018accelerating,xiao2019sharing}, the neural transformer has made remarkable progress in decoding speed, while preserves the translation performance. There are still 
great challenges when it is applied to production scenarios that require low-latency and high-throughput decoding. First of all, in the real-world commercial CLIR system, we need to deploy translation services around the world, but we don't have enough GPU machines in all computer rooms. Secondly, if the search query gets longer, the NMT decoding time will not meet the real-time performance requirements of the search.


\section{Related Work}

A lot of work has been tried to use the user clickthrough data of search engine to optimize a variaty of tasks in CLIR.
Hu et al. \cite{chen2013mining} proposed a method to use a dictionary for finding seed queries to mine URL pair patterns from the click-through data and then used the mined URL pair patterns to generate the candidate query pairs.  Ambati et al. \cite{ambati2006using} used a bilingual dictionary to translate queries in the monolingual click-through data of the target language and built a translation model from the resulting synthesized cross lingual query log to perform CLIR later. Gao et al. \cite{gao2007cross} mapped the input query of one language to queries of the other language in the query log by such as word translation relations and word co-occurrence statistics for query expansion task. Joachims \cite{joachims2002optimizing} utilized clickthrough data from the query log of the search engine in connection with the log of links the users clicked to train the ranking algorithm. Kreutzer et al. \cite{kreutzer2018can} presented methods to improve quality of title translation from human reinforcement signals which were logged from implicit task-based feedback collected in a cross-lingual search task. In contrast to previous studies, we purposefully mine user clickthrough data to uncover trustable bilingual pairs of multilingual queries and translations produced by a current query translation system of CLIR. 

Previous work mainly used bilingual dictionaries or SMT systems to perform query translation. However, most of these approaches require either expensive language resources or a large amount of parallel data, which are generally inconsistent with the domain of the queries. Pirkola et al. \cite{pirkola2001dictionary} were proposed to simplify the process of dictionary-based translation. Gao et al.  \cite{gao2001improving} enhanced the query translation by identifying the phrases using statistical model, then translating the phrases using set of phrase translation patterns and translating the remained words as words. Translating the queries using SMT system has shown potential improvement against other methods in CLIR. Sokolov et al.  \cite{sokolov2014learning} presented an method to optimise an SMT decoder to immediately output the best translation for CLIR. Bojar et al.  \cite{bojar2014findings} tuned SMT systems to produce the best translation in the perspective of humans. Wu et al.  \cite{wu2010study} conducted a set of CLIR experiments using free available translation system (eg. Google) for translating queries and showed that MT is an excellent tool for the query translation task.

To our knowledge, no such work has been done in studying the effect of NMT on query translation in CLIR, which has proven to be an effective technique in large-scale translation tasks in recent years. Bahdanau et al. \cite{bahdanau2014neural} proposed attention mechanism into encoder-decoder network and significantly improved the translation quality. The recently proposed Transformer \cite{vaswani2017attention} architecture achieves the best translation quality so far which relies on a self-attention mechanism \cite{lin2017structured} to calculate the representation of the source and target side sentences, removing all recurrent operations found in the former approach.

\section{CONCLUSION}
In this paper, we investigate the feasibility of exploiting NMT into CLIR.
We first empirically compare the effectiveness of SMT and NMT on query translation. Experimental results dispel the doubt on the inability of NMT on handling short query text. 
Then, we propose a novel approach to automatically acquiring a large amount of in-domain query translation pairs from the knowledge hidden in the user clickthrough data. The NMT system is able to be enhanced with the mined data using joint training strategy. Besides, to make the NMT technology practical in real-world commercial CLIR system, we devise a new system architecture that profits on both low latency of SMT and quality offered by NMT. 
To the best of our knowledge, this is the first work to apply the NMT technology to query translation for a real-world CLIR system. 



Our experiments show that NMT significantly improves CLIR performance over SMT and outperforms state-of-the-art CUNI system on CLEF 2018 multilingual task 4. Our proposed method is capable of collecting large-scale high-quality in-domain bilingual corpus from implicit user feedback. The extracted in-domain data are integrated into the normal training process of NMT through joint training and fine tuning. At the same time, the proposed architecture solves the problem of NMT online deployment in commercial system.

Although this work shows promising improvement in practice, there still exists several unsolved problems. First, the bilingual data extracted from the user behavior data contains noises which are harmful to the model training. Next, more quality estimation technologies can be explored to further filter out low-quality data. In addition, serving the user behavior as an indicator seems relatively simple, 
we will consider other indicators for data mining in future. Thirdly, the asynchronous strategy can be treated as a makeshift solution, more speed optimizations from algorithm and engineering perspectives are necessary.  
It is also interesting to integrate other improvements on NMT contexts~\cite{li2019neuron,wan2020self,zhou2020uncertainty} to further strengthen query translation.

%
\bibliographystyle{ACM-Reference-Format}
\bibliography{sample-base}

\end{document}